\documentclass[sigconf]{acmart}

\usepackage{multirow}
\usepackage{algorithm}
\usepackage{algpseudocode}
\usepackage{enumitem}

\usepackage{subcaption}
\usepackage{amsmath,amsfonts,bm}
\usepackage{booktabs}
\usepackage{array}

\usepackage{graphicx}

\newcommand{\figref}[1]{Figure~\ref{#1}}

\newcolumntype{L}[1]{>{\raggedright\let\newline\\\arraybackslash\hspace{0pt}}m{#1}}
\newcolumntype{C}[1]{>{\centering\let\newline\\\arraybackslash\hspace{0pt}}m{#1}}
\newcolumntype{R}[1]{>{\raggedleft\let\newline\\\arraybackslash\hspace{0pt}}m{#1}}

\DeclareMathOperator*{\argmin}{arg\,min}

\AtBeginDocument{%
  }

\setcopyright{acmlicensed}
\copyrightyear{2024}
\acmYear{2024}
\acmDOI{10.1145/3649329.3656528}

\acmConference[DAC '24]{61st ACM/IEEE Design Automation Conference}{June 23--27, 2024}{San Francisco, CA, USA}

\acmBooktitle{61st ACM/IEEE Design Automation Conference (DAC '24), June 23--27, 2024, San Francisco, CA, USA}

\acmISBN{979-8-4007-0601-1/24/06}

\begin{document}
\settopmatter{printacmref=false}
\title{Hardware-Aware Neural Dropout Search for \\ Reliable Uncertainty Prediction on FPGA}


\author{Zehuan Zhang}
\affiliation{%
  \institution{Department of Computing}
  \institution{Imperial College London}
  \country{}
}
\email{zehuanzhang22@imperial.ac.uk}
\author{Hongxiang Fan}
\affiliation{%
  \institution{Imperial College London \&\\Samsung Al Center}
  \country{}
}
\email{hongxiangfan@ieee.org}
\authornote{Corresponding author}

\author{Mark Chen}
\affiliation{%
  \institution{Department of Computing}
  \institution{Imperial College London}
  \country{}
}
\email{hao.chen20@imperial.ac.uk}
\author{Lukasz Dudziak}
\affiliation{%
  \institution{Samsung Al Center}
  \country{}
}
\email{l.dudziak@samsung.com}
\author{Wayne Luk}
\affiliation{%
  \institution{Department of Computing}
  \institution{Imperial College London}
  \country{}
}
\email{w.luk@imperial.ac.uk}

\renewcommand{\shortauthors}{Zhang et al.}
\renewcommand{\shorttitle}{Hardware-Aware Neural Dropout Search for Reliable Uncertainty Prediction on FPGA}

\begin{abstract}

The increasing deployment of artificial intelligence (AI) for critical decision-making amplifies the necessity for trustworthy AI, where uncertainty estimation plays a pivotal role in ensuring trustworthiness.
Dropout-based Bayesian Neural Networks (BayesNNs) are prominent in this field, offering reliable uncertainty estimates. 
Despite their effectiveness, existing dropout-based BayesNNs typically employ a uniform dropout design across different layers, leading to suboptimal performance. 
Moreover, as diverse applications require tailored dropout strategies for optimal performance,
manually optimizing dropout configurations for various applications is both error-prone and labor-intensive.
To address these challenges, 
this paper proposes a novel neural dropout search framework that automatically optimizes both the dropout-based BayesNNs and their hardware implementations on FPGA.
We leverage one-shot supernet training with an evolutionary algorithm for efficient dropout optimization.
A layer-wise dropout search space is introduced to enable the automatic design of dropout-based BayesNNs with heterogeneous dropout configurations.
Extensive experiments demonstrate that our proposed framework can effectively find design configurations on the Pareto frontier.
Compared to manually-designed dropout-based BayesNNs on GPU,
our search approach produces FPGA designs that can achieve up to 33$\times$ higher energy efficiency.
Compared to state-of-the-art FPGA designs of BayesNN,
the solutions from our approach can achieve higher algorithmic performance and energy efficiency.
\end{abstract}

\maketitle

\section{Introduction}
Deep neural networks (DNNs) are pervasively utilized in various domains~\cite{dong2021survey, fan2022optimizing}, achieving superior performance. 
Despite their capability, conventional DNNs are prone to overfitting, and thus are not able to indicate potential issues in their predictions.
This may cause silent failures for safety-critical applications~\cite{zou2023review}, compromising trustworthiness in deep learning for vital decision-making processes.
The capability of providing uncertainty estimation  is crucial in 
mitigating  the inherent risks in deep learning,
ensuring predictions come with well-calibrated confidence levels.
Bayesian Neural Network~\cite{neal1992bayesian} (BayesNN) emerges as a highly effective method for reliable uncertainty estimation.
Various approximation techniques have been introduced~\cite{jospin2022hands, hastings1970monte, blei2017variational} for BayesNN. Among these methods, dropout-based methods~\cite{gal2016dropout} have emerged as one of the mainstreaming approaches for reliable uncertainty estimation due to their compute and memory efficiency. Existing research has delved into exploring different dropout designs with diverse granularities and sampling
dynamics~\cite{ghiasi2018dropblock,zhang2019confidence}.

Although a substantial amount of progress has been made in this research field,
there are still challenges in deploying dropout-based BayesNNs in real-life applications. First, existing dropout-based BayesNNs are predominantly designed with a single dropout layer type throughout the network. This may lead to sub-optimal performance as different stages of the network require specialized dropout designs with distinct dropout granularities and sampling approaches to optimize performance. Second, diverse applications necessitate tailored dropout strategies to optimize performance. It is a labor-intensive and heuristic-driven process to manually find the optimal dropout configurations for the target application. Third, the computational and memory requirements of dropout-based BayesNNs necessitate hardware acceleration in practical deployment scenarios. However, previous accelerators~\cite{cai2018vibnn, awano2020bynqnet, fujiwara2021asbnn, fan2023monte,fan2022fpga,fan2022accelerating,fan2021high} are only designed for dropout-based BayesNN with a uniform dropout strategy. Also, the hardware efficiency of different dropout strategies is not considered while designing the dropout-based BayesNN accelerators.

To address the above challenges,
this paper proposes a novel neural dropout search framework that automatically optimizes both the dropout-based BayesNNs and the associated hardware accelerators on FPGA. 
The optimization of dropout strategies is formulated as a search problem with both algorithmic and hardware performance as the main objectives.
We develop a layer-wise dropout search space, 
allowing the hybrid use of different dropout layers at different network stages.
Our novel four-phase design framework automates the search of optimal dropout strategies, tailored to specific applications and constraints. The proposed framework leverages one-shot supernet training combined with an evolutionary algorithm to ensure efficient optimization.
Moreover, we introduce FPGA-based implementations for four distinct dropout designs, enabling the FPGA-based hardware acceleration of dropout-based BayesNNs with heterogeneous dropout layers. 
Our code is publicly available at: \url{https://github.com/zehuanzhang/Neural_Dropout_Search.git}.

Our contributions can be summarized as follows:
\begin{itemize}[leftmargin=*]
    \item A novel neural dropout search framework with one-shot supernet training and an evolutionary algorithm to automatically optimize both dropout-based BayesNNs and the associated FPGA-based accelerators given the target applications and constraints.

    \item A layer-wise dropout search space that enables the automatic optimization of dropout-based BayesNNs with heterogeneous dropout layers for higher performance.

    \item FPGA-based implementations of four types of dropout designs, enabling the acceleration of dropout-based BayesNNs with different dropout combinations.

\end{itemize}

\section{Background and Related Work}

\subsection{Background}
\subsubsection{Bayesian Neural Network}

In contrast to conventional DNNs,
BayesNNs employ probabilistic inference to provide uncertainty estimates for predictions.
A comprehensive literature review of BayesNNs is summarized in~\cite{jospin2022hands}.
Instead of presenting weights using point-wise values, BayesNNs are trained to construct the posterior distributions on weights through Bayesian inference.
Leveraging this approach, BayesNNs are able to deliver reliable uncertainty estimation, mitigating overfitting issues and enhancing robustness.

However, given the high dimensionality inherent in modern BayesNNs, the analytical inference of the posterior distribution associated with the model weights and the subsequent predictions is computationally prohibitive. To resolve this issue,
several approximation methods have been introduced~\cite{ovadia2019can, hastings1970monte, blei2017variational}, with dropout-based approximations showing significant promise.

\subsubsection{Dropout-based Approximation}\label{subsec:related_dropout}

Dropout-based approximations emerge as efficient approaches for approximating BayesNNs.
To obtain uncertainty estimation,
a dropout-based BayesNN runs the forward pass multiple times with the dropout enabled during inference.
Different dropout masks are generated in distinct forward passes to generate different Monte Carlo samples.

\begin{figure}[!t]
\centerline{\includegraphics[width=3.4in]{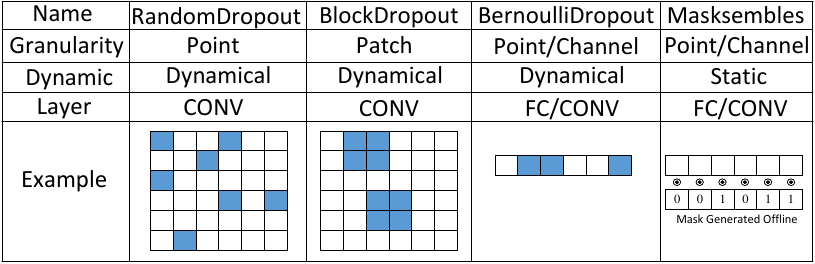}}
\caption{ Four dropout layers with different granularities and sampling dynamics.}
\label{fig:3_2}
\end{figure}

Employing dropout in both the training and inference stages of DNNs has demonstrated to perform Bayesian inference~\cite{gal2016dropout}.
Based on this theoretical grounding,
extensive studies have investigated various dropout designs with different granularities and sampling dynamics~\cite{ghiasi2018dropblock,zhang2019confidence}.
As shown in~\figref{fig:3_2},
this paper focuses on the four most representative dropout designs: Bernoulli ~\cite{gal2016dropout}, Block Dropout~\cite{ghiasi2018dropblock}, Random Dropout and Masksembles~\cite{durasov2021masksembles}.

\subsection{Related Work}

Given the substantial computational and memory demands, there has been significant research dedicated to accelerating BayesNNs.
\textit{VIBNN}~\cite{cai2018vibnn} accelerated BayesNNs based on Variational inference with two Gaussian (pseudo) random number generators employed for efficient inference.
\textit{BYNQNet}~\cite{awano2020bynqnet} explored the sampling-free method with an FPGA-based implementation on a Xilinx PYNQ-Z1 board.
\textit{ASBNN}~\cite{fujiwara2021asbnn} examined the relationship among multiple forward passes for approximate calculations.
To achieve higher hardware performance,
multiple studies~\cite{fan2021high,fan2022enabling,fan2022fpga} explored structured sparsity in different granularities.
\cite{fan2023monte} combined dropout methods and Multi-Exit networks to optimize dropout-based BayesNN with temporal and spatial mapping strategies.

These prior studies, however, predominantly involve manual designs of BayesNNs. Moreover, most approaches employing dropout-based BayesNNs neglect the vast potential offered by heterogeneous dropout layers.
In contrast to prior studies, this paper systematically analyzes different dropout strategies.
We propose a neural dropout search framework to find the optimal layer-wise dropout design, which is capable of automatically generating
an FPGA-based BayesNNs accelerator for the target application.

\section{Neural Dropout Search}\label{sec:framework}

\subsection{Framework Overview}

Given specifications and search objectives, our framework aims to find the optimal dropout configurations.
We formulate the optimization problem as follows:
\vspace*{-1.0ex}
\begin{equation}
\argmin\limits_{c \in C} L(c)
\label{eq:1}
\end{equation}

In the above, $C$ denotes the dropout search space, and $c$ denotes an individual dropout configuration. The overall training goal is to find the optimal dropout configuration scheme to meet algorithm and hardware requirements.

\begin{figure*}[ht]
\centerline{\includegraphics[width=7in]{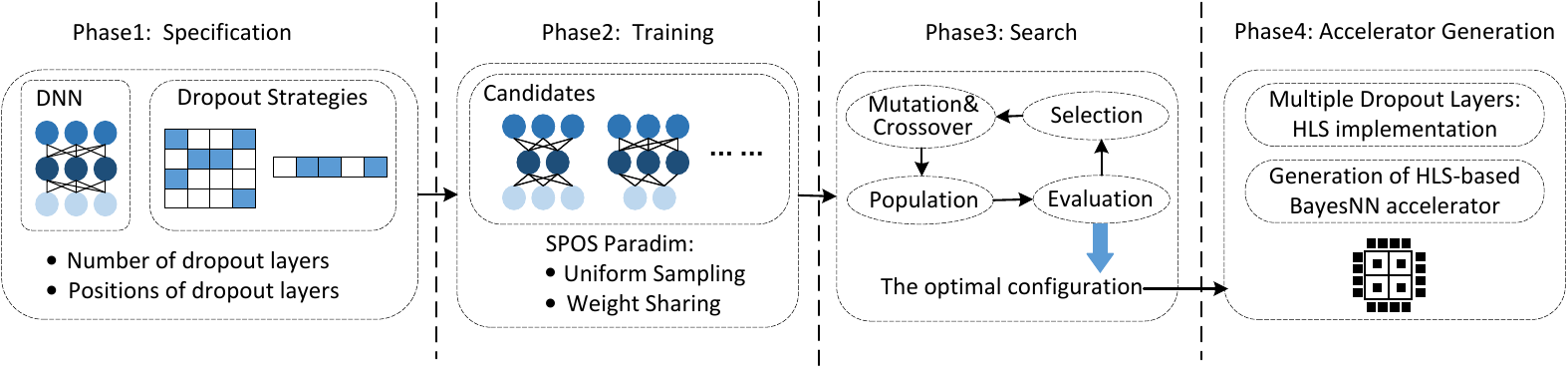}}
\vspace{-2mm}
\caption{ An overview of the proposed framework. HLS stands for High Level Synthesis.}
\label{fig:3_3}
\vspace{-1mm}
\end{figure*}

To address the optimization problem (\ref{eq:1}) above, we propose a novel neural dropout search framework as shown in Figure~\ref{fig:3_3}. To make the framework applicable to mainstream neural networks and dropout strategies, we systematically arrange it into four phases:      
(1)~Specification, (2)~Training, (3)~Search, (4)~Accelerator Generation.

\subsection{Specification: Phase1}

The framework starts by receiving the network architecture, heterogeneous dropout methods, and specified dropout layer positions as inputs. Typically, DNN architectures contain built-in dropout layers for regularization. Our framework is designed to support a wide variety of such DNNs. There are no restrictions imposed on network architectures or dropout strategies, so the proposed framework can support a wide spectrum of scenarios. 

A portion of built-in dropout layers is specified. Each specified dropout layer is assigned with multiple heterogeneous dropout methods, for constructing a supernet. The number of such specified dropout layers is denoted as $N$, and the number of dropout methods in each is denoted as $M_i (i = 1, 2, ..., N)$. For each forward pass, each specified dropout layer selects a dropout method randomly. Consequently, the constructed supernet encompasses an ensemble of network architectures, amounting to a total of $\prod_{i=1}^{N} M_i$ sub-networks characterized by diverse configurations, each serving as a candidate within the total search space. Note that candidates with uniform and hybrid dropout configurations are both allowed considerably expanding the exploration space.

Regarding the heterogeneous dropout methods, four dropout designs—Random Dropout, Block Dropout~\cite{ghiasi2018dropblock}, Bernoulli Dropout~\cite{gal2016dropout} and Masksembles~\cite{durasov2021masksembles}—are employed. 
The first three dropout methods support dynamic adoption with different levels of granularity, while the last one is static and can be placed following both convolutional layers and fully connected layers. Different dropout designs offer a means to strike a balance between accuracy, uncertainty estimation, and hardware performance.

\subsection{Training: Phase2}

As stated in Eq (\ref{eq:1}), the training process can be cast as a multi-objective problem. A naive approach to solve this is to directly train the supernet from scratch using the gradients of the overall objective derived by evaluating all sub-networks at each update step. However, this approach incurs an increase in training costs proportional to the number of sub-networks. If the size of the search space is enormous, it is challenging to train every candidate sub-network. Moreover, when handling different tasks, it necessitates retraining the supernet, which can be computationally prohibitive. 

Inspired by \cite{guo2020single}, the Single Path One-Shot (SPOS) paradigm is exploited for  supernet training. Within each training iteration, a candidate sub-network is uniformly sampled by randomly selecting the dropout strategy in each specified layer. This paradigm allows for the sharing of weights across all sub-networks, thus enabling convergence of each candidate sub-network even if the search space is enormous. As such, it results in substantial reduction of the overall training costs from $O(\prod_{i=1}^{N} M_i)$ to $O(1)$, significantly addressing the computational challenges posed. Furthermore, the training and searching stages are decoupled. The supernet needs to be trained only once. Subsequently, each candidate sub-network can be directly sampled with shared weights in the search phase.

\begin{figure}[ht]

\centerline{\includegraphics[width=3.4in]{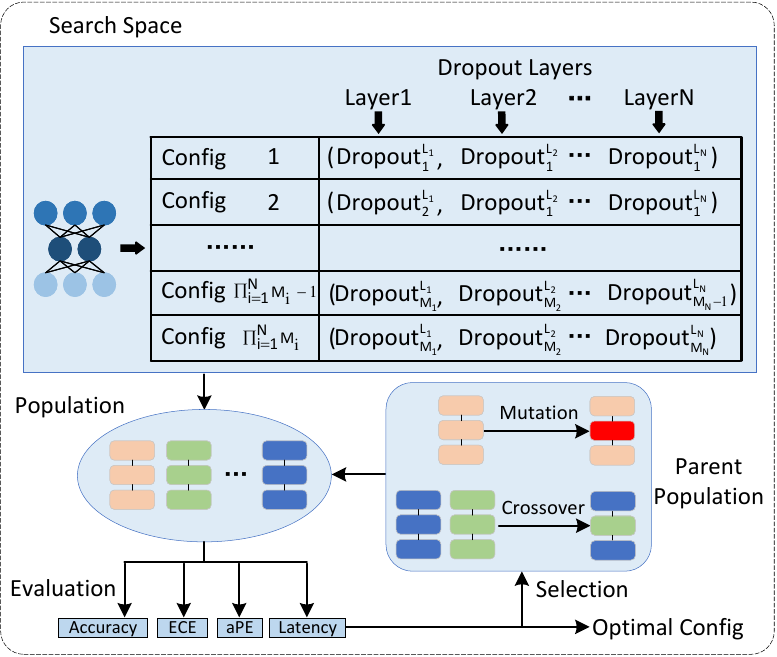}}
\caption{ Evolutionary Algorithm. }
\label{fig:3_4}
\vspace{-5mm}
\end{figure}

\subsection{Search: Phase3}

After training, the third phase is to conduct a thorough search for the optimal network configuration given the defined search aim, which takes into account both software and hardware performance. An evolutionary algorithm is adopted to search for the optimal configuration as shown in Figure~\ref{fig:3_4}.

Our evolutionary algorithm consists of four stages: population, evolution, selection, and crossover\&mutation. 
First, the population stage initiates by randomly sampling a set of candidate sub-networks with distinct dropout configurations. Each candidate sub-network can be represented as a combination of dropout designs. The sampling process continues until the candidate pool reaches the predefined size. 
Second, the performance of all the sampled candidates in the population is evaluated on the validation set. 
Third, the top-performing solutions are selected as the parent population. 
Fourth, a fraction of candidates in the parent population undergo mutations, while the others undergo crossovers. 
During the mutation process, each candidate experiences random mutation in the dropout types with a predefined probability. 
During the crossover process, a pair of candidates are selected randomly. A new configuration is produced by swapping each dropout type randomly within the specified dropout layers. Through the crossover and mutation processes, new network configurations are generated to update the population candidates. 
Such sampling and updating processes are repeated until the optimal configuration is found.

Within our framework, we choose the accuracy metric to assess the predicted performance, the expected calibration error (ECE) and average prediction entropy (aPE) metrics to assess the uncertainty performance, and hardware latency to measure the running speed for each potential candidate. The algorithmic metrics are evaluated on the validation set.
The search aim is expressed as:
\begin{equation}
aim = \eta \times Accuracy - \mu \times ECE + \beta \times aPE -  \lambda \times Latency
\label{eq:3_5}
\end{equation}

Here $\eta$, $\mu$, $\beta$ and $\lambda$ represent the weights of the respective metrics in the search. As the lower the ECE and latency are, the higher the overall performance is, so these two metrics are negative terms. The final goal is to find a network architecture that maximizes both accuracy and uncertainty performance while minimizing hardware latency. 
As the metrics cover trade-offs that need to be considered and balanced, it is allowed to prioritize these metrics and establish the search aim accordingly tailored to specific requirements.

\subsection{Accelerator Generation: Phase4}

\subsubsection{Performance modelling}
As explained above,
the third phase of our framework employs an evolutionary algorithm to optimize both algorithmic and hardware settings. Each iteration of the evolutionary algorithm necessitates running the hardware implementation to assess the performance of the chosen candidates for subsequent optimization.
Given that design synthesis and place \& route operations in FPGA hardware implementation are time-consuming and often need manual intervention, the optimization process can become lengthy. To expedite this process, we introduce a machine learning-based hardware cost model, facilitating rapid evaluation.

For accurate and fast estimation,
we construct a training dataset that consists of multiple data points, where hardware configurations serve as inputs and predicted values of latency as the output.
The hardware configurations include the input shape and dropout type.
We employ Gaussian process~\cite{williams2006gaussian} for regression using this dataset.
It is worth noting that both dataset construction and training are only required once.
The Gaussian process model can be reused for various evolutionary optimizations with different search objectives.
We choose Matérn kernel and constant mean function.

\subsubsection{HLS-based implementation}\label{4_1}
The last phase generates the corresponding hardware accelerator using HLS (High Level Synthesis). The HLS4ML\cite{fahim2021hls4ml} open-source framework is adopted to generate our BayesNNs accelerators.
There are existing HLS templates of normal layers in HL4ML to construct traditional DNNs. To support BayesNN accelerators, we introduce HLS-based implementation of the newly introduced dropout layers into the design flow. 
So hardware accelerators, designed using HLS, can then be integrated into Vivado-HLS for synthesis and place $\&$ route.

\section{Experiments}

Experiments are implemented in Python 3.9, Pytorch 1.10.0, and Keras 2.9.0. The Intel Core i9-9900K CPU and one Nvidia GeForce GTX 2080 TI GPU are equipped in our machine. We use Vivado-HLS 2020.1 for hardware implementation. 16-bit fixed data is used, with 1 sign bit, 7 integer bits and 8 fraction bits. QKeras is used for quantization. The latency, resource utilization and power consumption are obtained from C-synthesis reports provided by Vivado-HLS. Vivado 2020.1 is used to run place and route for the final designs. We select Xilinx Kintex XCKU115 as our target FPGA board.

\begin{table*}[]
\caption{Algorithm and hardware results of optimized configurations obtained from search.} 
\label{tab:5_1} 
\begin{tabular}{l|p{3.7cm}|p{1.4cm}|p{1.4cm}|p{1.4cm}|p{1.4cm}|lll}
\hline
\multirow{2}{*}{\textbf{ResNet}}                                                            & \multirow{2}{*}{\textbf{Configurations}} & \multirow{2}{*}{\textbf{\begin{tabular}[c]{@{}l@{}}Accuracy\\ (\%)$\uparrow$\end{tabular}}} & \multirow{2}{*}{\textbf{\begin{tabular}[c]{@{}l@{}}ECE\\ (\%)$\downarrow$\end{tabular}}} & \multirow{2}{*}{\textbf{\begin{tabular}[c]{@{}l@{}}aPE\\ (nats)$\uparrow$\end{tabular}}} & \multirow{2}{*}{\textbf{\begin{tabular}[c]{@{}l@{}}Latency\\ (ms)$\downarrow$\end{tabular}}} & \multicolumn{3}{l}{\textbf{Resource Utilization}}     \\ \cline{7-9} &    &      &     &     &     & \multicolumn{1}{l|}{\textbf{BRAM}} & \multicolumn{1}{l|}{\textbf{DSP}} & \multicolumn{1}{l}{\textbf{FF}} \\ \hline
\multirow{4}{*}{\textbf{\begin{tabular}[c]{@{}l@{}}Uniform\\ Configurations\end{tabular}}}  & \textbf{All Bernoulli Dropout}           & 91.205     & 7.4  & 0.989    & 15.401   & \multicolumn{1}{l|}{82\%}          & \multicolumn{1}{l|}{5\%}          & \multicolumn{1}{l}{40\%}              \\ \cline{2-9} & \textbf{All Block Dropout}  & 91.276    & 5.9    & 0.887    & 18.674    & \multicolumn{1}{l|}{82\%}          & \multicolumn{1}{l|}{5\%}          & \multicolumn{1}{l}{39\%}               \\ \cline{2-9} & \textbf{All Random Dropout}              & 90.635    & 5.8  & 0.773   & 18.396   & \multicolumn{1}{l|}{82\%}          & \multicolumn{1}{l|}{5\%}          & \multicolumn{1}{l}{39\%}             \\ \cline{2-9} & \textbf{All Masksembles}                 & 91.316     & 3.6   & 0.626     & 15.401      & \multicolumn{1}{l|}{82\%}          & \multicolumn{1}{l|}{5\%}          & \multicolumn{1}{l}{39\%}               \\ \hline
\multirow{4}{*}{\textbf{\begin{tabular}[c]{@{}l@{}}Searched\\ Configurations\end{tabular}}} & \textbf{Accuracy Optimal}                      & \textbf{91.456}    & 5.5   & 0.784    &  18.671    & \multicolumn{1}{l|}{82\%}          & \multicolumn{1}{l|}{5\%}          & \multicolumn{1}{l}{40\%}               \\ \cline{2-9} & \textbf{ECE Optimal}                      & 91.316    & \textbf{3.6}   & 0.626    & 15.401    & \multicolumn{1}{l|}{82\%}          & \multicolumn{1}{l|}{5\%}          & \multicolumn{1}{l}{39\%}               \\ \cline{2-9} & \textbf{aPE Optimal}                      & 91.205      & 7.4    & \textbf{0.989}     & 15.401     & \multicolumn{1}{l|}{82\%}          & \multicolumn{1}{l|}{5\%}          & \multicolumn{1}{l}{40\%}              \\ \cline{2-9} & \textbf{Latency Optimal}                  & 91.206      & 7.4    & 0.626    & \textbf{15.401}     & \multicolumn{1}{l|}{82\%}          & \multicolumn{1}{l|}{5\%}          & \multicolumn{1}{l}{39\%}              \\ \hline
\end{tabular}
\end{table*}

\subsection{Effectiveness of Neural Dropout Search}

To demonstrate the advantages of auto-generated configurations of dropout-based BayesNNs, we evaluate three commonly-used models, LeNet, VGG11 and ResNet18 for image classification on different datasets, MNIST, SVHN and Cifar-10, respectively.  
Synthetic data generated by random Gaussian noise with mean and standard deviation of the training data, are employed to measure aPE.
There are four dropout choices: Bernoulli Dropout, Random Dropout, Block Dropout and Masksembles. For LeNet, we specified three dropout layers: (a)~two dropout layers follow convolutional layers with all four dropout choices, (b)~one dropout layer follows fully-connected layers with two dropout choices: Bernoulli Dropout and Masksembles. For VGG11 and ResNet18, we specify four dropout layers following convolutional layers with four dropout choices. The sampling number is set as three.

\begin{figure}[ht]
\ \\
\centerline{\includegraphics[width=3.3in]{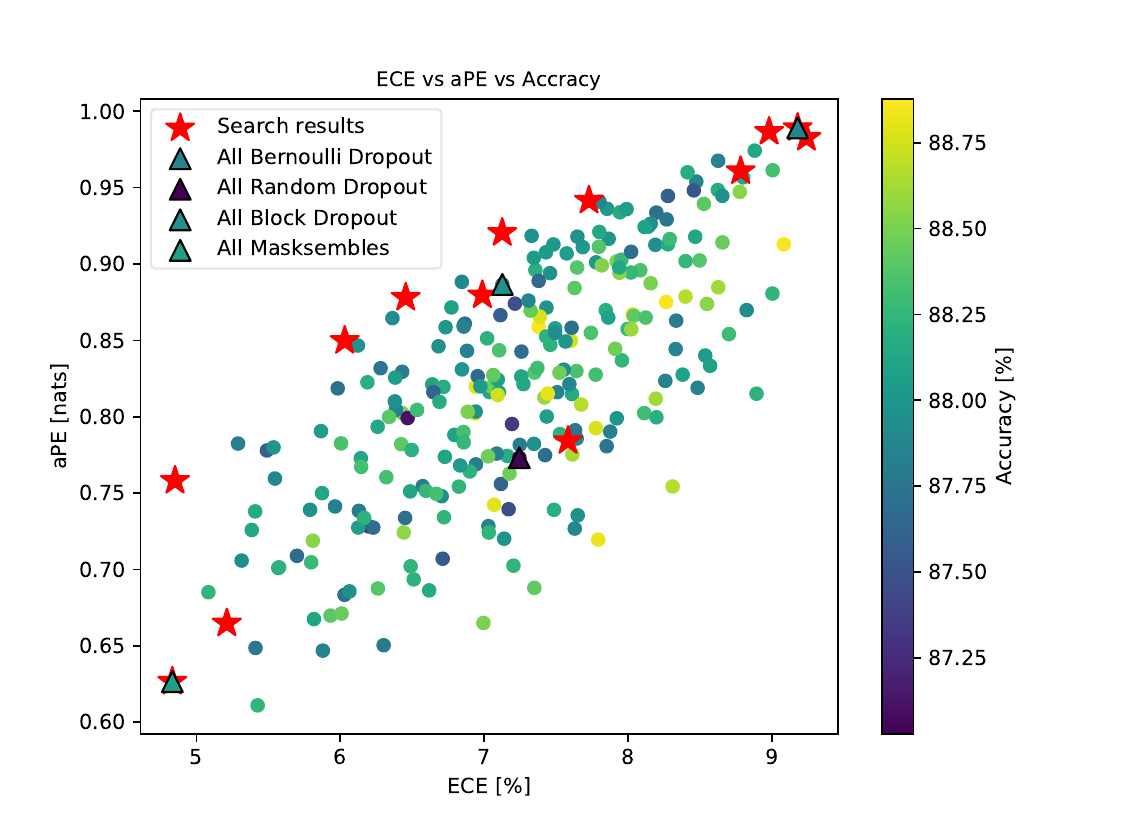}}
\caption{Search results.}
\label{fig:5_1c}
\end{figure}

The results of ResNet18 on the test set of Cifar-10 are presented in Table \ref{tab:5_1}. We adopt accuracy, ECE, aPE and latency as search aims, respectively. Table~\ref{tab:5_1} shows that all the optimal configurations can be found. 
To further demonstrate the effectiveness of our search approach, we adjust search aims to find different Pareto-optimal designs. The Pareto-optimal points outperform any other points in either accuracy or uncertainty.
As discussed, the weight parameters in the search aim represent the importance of different metrics. We iterate through and evaluate all configurations on the validation sets.
To compare in a straightforward way, we highlight baselines configured with uniform dropout designs and the configurations obtained from search.
As shown in Figure~\ref{fig:5_1c}, all the searched results lie on the reference Pareto frontier, confirming that our framework can effectively identify optimal designs.
Therefore, for real applications, multiple configurations can be obtained depending on different aims, providing strong flexibility.

The search costs and the resulting configurations are shown in Table \ref{tab:search_config}. To achieve the highest accuracy, the optimal dropout configurations for LeNet, VGG11, and  ResNet18 are all hybrid dropout configurations. In previous work, the networks are typically designed with uniform dropout configurations manually. Hybrid dropout configurations are rarely designed by human experts, leading to oversight of optimal designs.
Moreover, in some cases, the uniform configurations may achieve performance even below the average level, such as the `All Random Dropout' in ResNet in Figure~\ref{fig:5_1c}. 
This further emphasizes the benefits of our approach. Given a vast design space and the variability of tasks, our auto-search framework can effectively uncover configurations that may be non-intuitive to human experts yet yield superior performance outcomes.

\subsection{Comparison with CPU and GPU}
We compare our approach against CPU and GPU implementations as shown in Table~\ref{tab:5_2}. The comparison uses MNIST dataset since it is the most common dataset in prior work. For dropout-based BayesNNs, the Bernoulli Dropout method is adopted as it is widely employed in hand-crafted approaches. We implement the auto-search configurations targeting the aPE Optimal design.

Regarding aPE performance, our auto-search design outperforms the model configured with a uniform Bernoulli Dropout. In terms of hardware performance, our design achieves 1.4 times speedup over CPU implementation. The power consumption of our design is $52.6\times$ and $60.5\times$ lower than that of 
CPU and GPU implementations. 
Consequently, our design achieves $65\times$ and $33\times$  higher energy efficiency over CPU and GPU implementations, despite the more advanced 14nm technology used in CPU and 12nm in GPU.



\begin{table}[htb]
\centering
\caption{Search costs and resultant configurations on three networks. (B: Bernoulli Dropout, R: Random Dropout, K: Block Dropout, M: Masksembles)}
\vspace{1.0 mm}
\label{tb:resource_breakdown}
\setlength\tabcolsep{1pt} 
\scalebox{0.9}{
\begin{tabular}{C{1.5cm}|C{3.0cm}|C{4.5cm}}
\toprule
& \textbf{Search Cost (GPU)} & \textbf{Dropout Configurations}  \\ \midrule
\multirow{4}{*}{\textbf{LeNet}} & \multirow{4}{*}{$\thicksim$2 hours}  & Accuracy Optimal: B - B - M  \\ \cmidrule{3-3}
&& ECE Optimal: M - M - B  \\ \cmidrule{3-3}
&& aPE Optimal: R - R - B \\ \cmidrule{3-3}
&& Latency Optimal: M- M- M  \\ \midrule
\multirow{4}{*}{\textbf{VGG}} & \multirow{4}{*}{$\thicksim$6 hours}  & Accuracy Optimal: R - B - B - R  \\ \cmidrule{3-3}
&& ECE Optimal: R - K - R - M  \\ \cmidrule{3-3}
&& aPE Optimal: R - R - R - R \\ \cmidrule{3-3}
&& Latency Optimal: M- M- M -M  \\ \midrule
\multirow{4}{*}{\textbf{ResNet}} & \multirow{4}{*}{$\thicksim$10 hours}  & Accuracy Optimal: K - M - B - M  \\ \cmidrule{3-3}
&& ECE Optimal: M - M - M - M  \\ \cmidrule{3-3}
&& aPE Optimal: B - B - B - B \\ \cmidrule{3-3}
&& Latency Optimal: M- M- M -M  \\ \midrule
\end{tabular}}
\vspace{-1.5mm}
\label{tab:search_config}
\end{table}

\begin{table*}[t]
\caption{Comparisons with related work.} 
\label{tab:5_2} 
\begin{tabular}{p{4cm}p{2cm}p{2cm}llp{2cm}lp{2cm}lp{2cm}lp{2cm}lp{2cm}lp{2cm}l}
\hline
-      & \textbf{CPU} & \textbf{GPU} & \textbf{ASPLOS'18 }\cite{cai2018vibnn} & \textbf{DATE'20 }\cite{awano2020bynqnet} & \textbf{TPDS'22 }\cite{fan2022accelerating}  & \textbf{Our Work} 
\\ \hline
\textbf{Platform}                  & \begin{tabular}[c]{@{}l@{}}Intel Core\\ i9-9900K\end{tabular}              & \begin{tabular}[c]{@{}l@{}}NVIDIA\\ RTX 2080\end{tabular}              & \begin{tabular}[c]{@{}l@{}}Altera\\  Cyclone V\end{tabular}    & \begin{tabular}[c]{@{}l@{}}Zynq\\  XC7Z020\end{tabular}      & \begin{tabular}[c]{@{}l@{}}Arria 10\\ GX1150\end{tabular}     & XCKU115                 
\\ \hline
\textbf{Frequency(MHz)}             & 3600            & 1545            & 213                & 200              & 220             & 181                 
\\ \hline
\textbf{Technology}                 & 14 nm            & 12 nm            & 28nm               & 28nm             & 20nm           & 20nm                 
\\ \hline
\textbf{Power(W)}                   & 205            & 236            & 6.11               & 2.76             & 43.6           & 3.9                 
\\ \hline

\textbf{aPE(nats)$\uparrow$}                 & 0.27           & 0.27            & -              & -            & 0.45              &  \textbf{0.65}           
\\ \hline
\textbf{Latency(ms)}                & 1.26            & 0.57            & 5.5                & 4.5              & 0.32               & 0.905                 \\ \hline
\textbf{Energy Efficiency(J/Image)$\downarrow$} & 0.258            & 0.134            & 0.033              & 0.012            & 0.014               & \textbf{0.004}                \\ \hline
\end{tabular}
\end{table*}

\subsection{Comparison with Related Work}

To compare with related work, we quote results from relevant papers. The results are shown in Table~\ref{tab:5_2}. Both \cite{cai2018vibnn} and \cite{awano2020bynqnet} do not support LeNet. In terms of hardware performance, our accelerator achieves 6.1 times and 5.0 times faster speed than \cite{cai2018vibnn} and \cite{awano2020bynqnet}, and shows 8.2 times and 3.0 times higher energy efficiency than those, respectively. 
To compare with the work in~\cite{fan2022accelerating}, we reproduce the experiments using its techniques with the same sampling number as ours to report the aPE metric. The hardware performance is quoted from~\cite{fan2022accelerating}. Our design achieves better aPE than~\cite{fan2022accelerating}, demonstrating the advantages of our auto-search framework. Although it operates at a lower latency, it incurs higher power costs, resulting in lower energy efficiency than our work.
Since the design in~\cite{fan2023monte} adopts multi-exit optimization to enhance performance, we are not able to compare against it.
As their techniques are orthogonal to ours,
we intend to incorporate these optimizations in future.



\begin{figure}[t!]
    \begin{minipage}[t]{0.5\linewidth}
        \centering
        \includegraphics[width=1.0\textwidth]{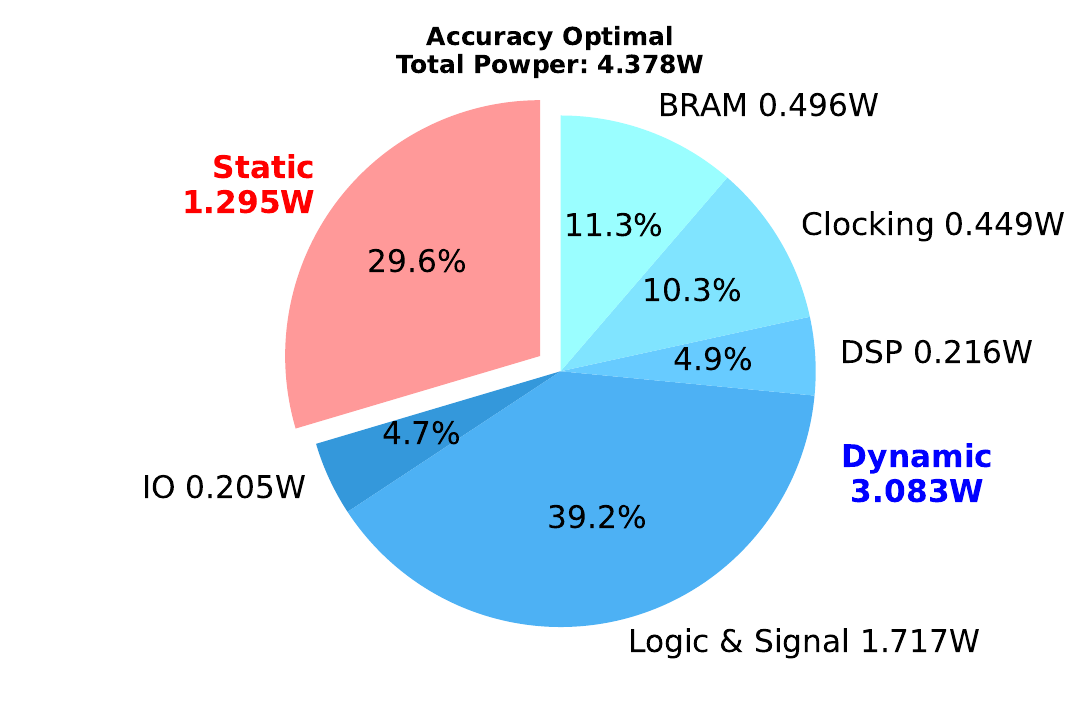}
    \end{minipage}%
    \begin{minipage}[t]{0.5\linewidth}
        \centering
        \includegraphics[width=1.0\textwidth]{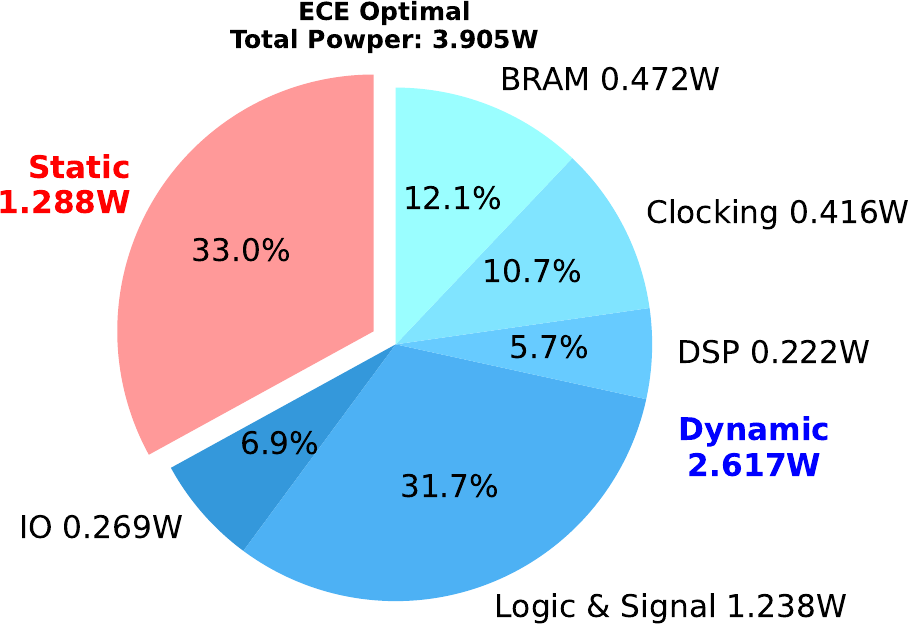}
    \end{minipage}
    \caption{ Power breakdown of the search designs. }
    \label{fig:4_3}
\end{figure}

Figure~\ref{fig:4_3} presents the power consumption breakdown of Accuracy Optimal and ECE Optimal designs, estimated from the Vivado tool after place and route. The Logic\&Signal accounts for $39\%$ and $32\%$ of the dynamic power, respectively. The high consumption is due to the comparing operations in dynamic dropout layers. The BRAM accounts for $11\%$ and $12\%$ of the dynamic power. The implementation of Masksembles consumes more BRAM resources.


In addition, our design is more versatile and flexible than existing designs. The accelerators~\cite{cai2018vibnn,awano2020bynqnet} only support BayesNNs consisting of fully-connected layers. Our design is not specific to a particular network, and more dropout types are accepted. The framework can also be extended to cover more deep learning architectures, providing high generality. Furthermore, our accelerator is designed using HLS, which offers the benefits of improved productivity and easy code portability compared with RTL design~\cite{fan2022accelerating}.

\section{Conclusion}

This paper proposes a novel neural dropout search framework for automatic optimization of dropout-based BayesNNs and their implementation as hardware accelerators targeting FPGA technology.
We leverage one-shot supernet training and an evolutionary algorithm approach to search for the optimal dropout designs.
A layer-wise dropout search space is introduced to enable hybrid dropout design for BayesNNs.
We propose hardware implementation for four different dropout layers, allowing the efficient mapping of the optimized BayesNNs onto an FPGA device.

Extensive experiments demonstrate that our approach can effectively identify the Pareto frontier designs with higher performance than prior methods.
Further research includes incorporating additional dropout designs into our search space, providing sparsity support for hardware design, and extending the proposed framework to cover other kinds of neural networks such as Transformer and Graph Neural Network.

\section*{Acknowledgement}{
The support of UK EPSRC (grant number EP/X036006/1, EP/P010040/1, EP/V028251/1 and EP/S030069/1) is gratefully acknowledged.
}



\bibliographystyle{ACM-Reference-Format}
\bibliography{_reference}

\end{document}